\documentclass[10pt,twocolumn,letterpaper]{article}
\usepackage[pagenumbers]{cvpr}

\usepackage{placeins}

\definecolor{cvprblue}{rgb}{0.21,0.49,0.74}
\usepackage[pagebackref,breaklinks,colorlinks,allcolors=cvprblue]{hyperref}

\def\confName{CVPR}
\def\confYear{2027}
\newcommand{\method}{Admoab-DETR}
\newcommand{\rgb}{\mathrm{V}}
\newcommand{\ir}{\mathrm{I}}
\newcommand{\sg}{\operatorname{sg}}
\newcommand{\cat}{\operatorname{Cat}}

\title{Admoab-DETR: Resolution-Tiered Fusion for Infrared–Visible Object Detection with Missing Modalities}

\author{%
  Wenchao Li\textsuperscript{1}\quad
  Haoran Jia\textsuperscript{2}\quad
  Shuyuan Wen\textsuperscript{3}\quad
  Keyan Chen\textsuperscript{4}\\[0.5em]
  \normalsize\textsuperscript{1}Department of Automation, Tsinghua University\\
  \normalsize\textsuperscript{2}School of Mathematics, Peking University\\
  \normalsize\textsuperscript{3}College of Control Science and Engineering, China University of Petroleum (East China)\\
  \normalsize\textsuperscript{4}Nanyang Technological University
}
\hypersetup{pdfauthor={Wenchao Li, Haoran Jia, Shuyuan Wen, Keyan Chen}}

\begin{document}
\maketitle
\begin{abstract}
Infrared and visible images offer complementary cues for all-weather object detection, yet real-world sensor dropouts can leave one stream unavailable and degrade paired-input detectors. Meanwhile, existing modality-adaptive methods apply heavy cross-modal attention uniformly across scales, incurring prohibitive token overhead on high-resolution maps. We propose Admoab-DETR, a unified real-time detector supporting infrared-only, visible-only, and paired inputs with a single set of weights. Specifically, \emph{Resolution-Tiered Fusion} (RTF) allocates fusion capacity by spatial token density and feature granularity: a lightweight Spatial Modality Gate governs fine local fusion at the highest-resolution scale with zero-cost single-view pass-through, while channel-compressed deformable Admoab necks capture cross-modal correspondence at coarser semantic scales. Furthermore, \emph{Mode-Matched Feature Distillation} (MMFD) aligns multi-scale encoder features with an availability-synchronized full-fusion teacher, avoiding ill-posed cross-modal hallucination while cutting parameters by 18.1\% and paired latency by 39.6\% at zero inference overhead. We also construct Admoab, a strictly registered 20,000-pair infrared--visible benchmark across diverse scenes. Experiments show Admoab-DETR achieves state-of-the-art accuracy--efficiency trade-offs across missing-modality and paired benchmarks. Our code will be available at https://github.com/liwenchao0615/Admoab-DETR.
\end{abstract}

\section{Introduction}
\label{sec:intro}

\begin{figure*}[t]
\centering
\includegraphics[width=\textwidth]{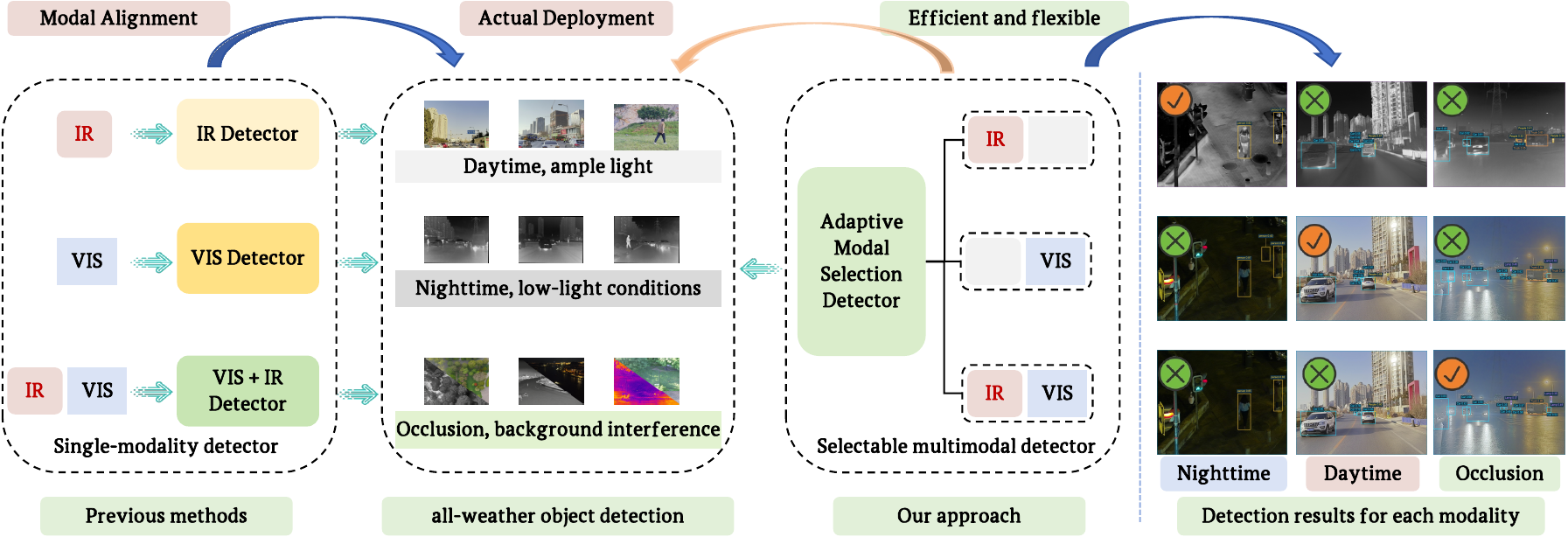}
\caption{\textbf{Object Detection with Adaptive Modal Selection.} Unlike conventional single- or fixed-modality detectors, our unified detector adapts to available modalities across diverse lighting and occlusion conditions to facilitate practical deployment.}
\label{fig:architecture}
\end{figure*}

In clear weather with sufficient illumination, visible-light images provide rich texture and appearance cues. Under darkness, visible details degrade sharply, whereas thermal infrared imaging captures emitted radiation differences between targets and backgrounds. In fog, heavy clutter, or partial occlusion, the two modalities provide complementary cues for discerning targets ambiguous in either view alone, motivating visible--thermal fusion for all-weather object detection.

Existing detectors span visible-only, infrared-only, and dual-modality configurations. Single-modality detectors---such as the YOLO~\cite{yoloOriginal,yoloEleven} and DETR/RT-DETR~\cite{detr,rtDetr,rtdetrv4} families---perform well in favorable conditions, yet remain bounded by single-sensor physics and suffer from misses or false alarms under darkness, low thermal contrast, distant small objects, or clutter. Scaling single-modality networks raises parameters and latency without overcoming sensor blind spots.

Multi-modal detectors leverage complementary cues for complex scenes: DAMSDet~\cite{damsDet} guides cross-modal sampling with object queries, and GM-DETR~\cite{gmDetr} pairs two-stage single-modal pretraining with fusion training. In practical deployments, however, sensor dropouts or bandwidth limits can leave one view unavailable, degrading models assuming complete pairs. Although Scarf-DETR~\cite{scarfDetr} supports single- and dual-modal inputs via shared deformable fusion, applying identical heavy blocks across all scales overlooks a key structural asymmetry: the finest map ($P_3$) holds over $76\%$ of pyramid tokens while mainly encoding local edges rather than long-range cross-modal shifts, creating a severe latency bottleneck. Moreover, conventional cross-modal distillation forces missing-modality students to hallucinate absent paired features, and public benchmarks rarely offer large-scale, strictly registered pairs across diverse scenes.

To address these challenges, we propose Admoab-DETR (Figure~\ref{fig:architecture}), a unified real-time detector supporting visible-only, infrared-only, and paired inputs within one set of weights. Departing from uniform fusion, Admoab-DETR introduces \emph{Resolution-Tiered Fusion} (RTF) matched to cross-scale token density and granularity: at token-dense $P_3$, a position-wise Spatial Modality Gate (SMG) weights local paired cues and reduces to a zero-overhead pass-through for single views; at coarser $P_4$ and $P_5$, channel-compressed deformable Admoab necks with uncompressed residual anchoring preserve cross-modal spatial sampling at minimal cost. We further propose \emph{Mode-Matched Feature Distillation} (MMFD), synchronizing modality availability between a frozen full-fusion teacher and the student across all encoder scales to transfer capacity without cross-modal hallucination or deployment overhead. Finally, we construct Admoab, comprising approximately 20,000 strictly registered infrared--visible pairs across day/night, small-target, occluded, and cluttered scenes verified for geometric consistency.

Our main contributions are summarized as follows:
\begin{itemize}
\item \textbf{Asymmetric Scale-Aware Architecture:} We propose Admoab-DETR with Resolution-Tiered Fusion (RTF), coupling fine-scale spatial gating with coarse-scale compressed deformable aggregation to support paired and single-view detection in real time within one checkpoint.
\item \textbf{Availability-Consistent Distillation:} We introduce Mode-Matched Feature Distillation (MMFD), aligning multi-scale features under matched modality availability to eliminate cross-modal hallucination at zero inference cost.
\item \textbf{Large-Scale Registered Benchmark:} We construct Admoab, a strictly aligned 20,000-pair visible--infrared dataset across diverse scenes, demonstrating consistent detection gains from geometric registration.
\item \textbf{State-of-the-Art Accuracy and Inference Speed: } Admoab-DETR sets a new state of the art in both detection accuracy and inference speed, consistently outperforming existing paired- and missing-modality detectors across all input modes.
\end{itemize}

\begin{figure*}[t]
\centering
\includegraphics[width=\textwidth]{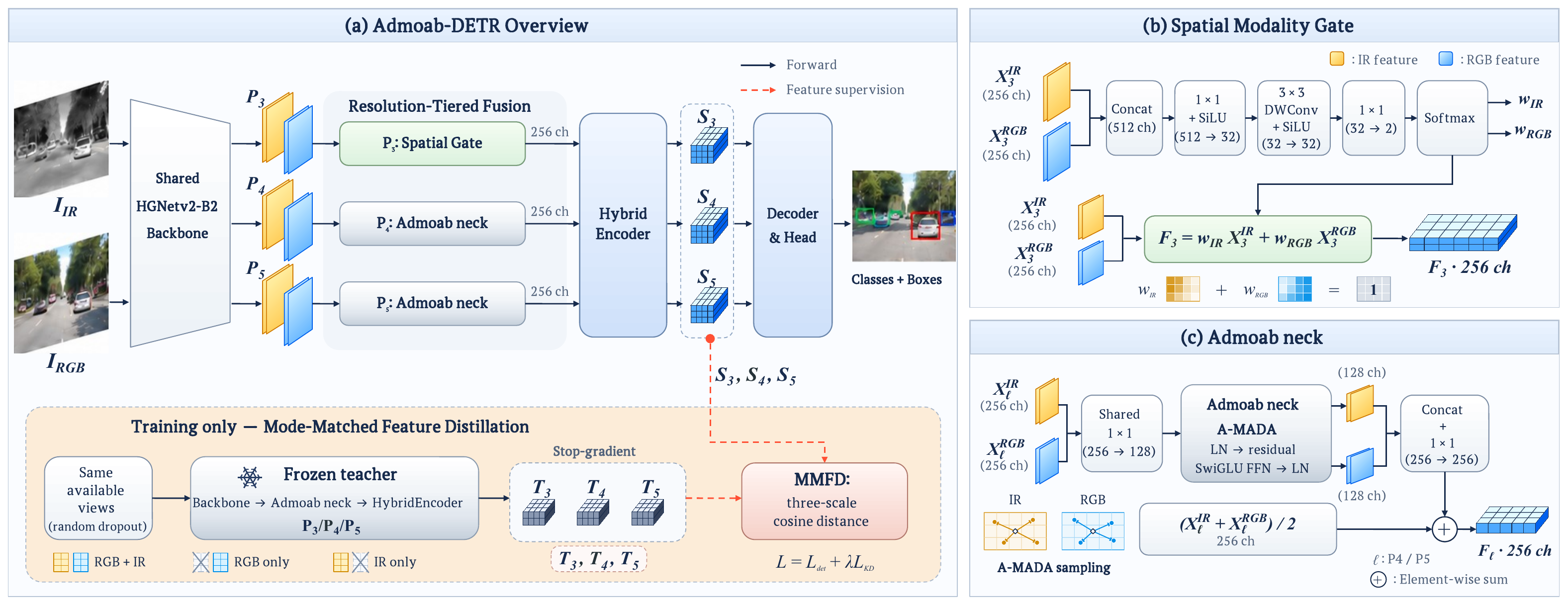}
\caption{\textbf{Architecture of \method{}.} (a) Unified modality-routed detector with Resolution-Tiered Fusion (RTF) and training-only Mode-Matched Feature Distillation (MMFD). (b) Finest-scale P3 Spatial Modality Gate with single-view pass-through. (c) Compact P4/P5 Admoab necks with bottleneck deformable sampling and modality-consistent residual anchoring.}
\label{fig:method_architecture}
\end{figure*}

\section{Related Work}
\label{sec:related}

\subsection{Thermal Infrared Object Detection}
Thermal infrared sensors capture emitted long-wave radiation, enabling target perception in complete darkness, smoke, and glare. To handle weak textures and tiny signatures, specialized infrared architectures aggregate multi-scale context via asymmetric modulation~\cite{acmInfrared}, dense nested attention~\cite{dnanetInfrared}, or joint super-resolution and detection~\cite{thermalSrDetection}. Because labeled thermal datasets are scarcer than RGB benchmarks, cross-domain transfer adapts visible-light detectors through pseudo multi-modal generation~\cite{borrowThermal}, day--night task conditioning~\cite{taskConditionedThermal}, self-supervised contrastive alignment~\cite{sstnThermal}, and meta-learned domain adaptation~\cite{metaUdaThermal}. Recent methods further transfer multimodal priors into standalone thermal detectors via decomposed distillation~\cite{decomKdThermal} or pseudo-visible feature synthesis~\cite{pfgfThermal}. Yet thermal-only streams lack fine appearance details under low thermal contrast, and distilling paired teachers into a single-view student forces ill-posed hallucination of absent visible cues.

\subsection{Visible-Light Object Detection}
Visible-light detection has evolved across multi-scale feature pyramids~\cite{fpn}, two-stage region-based cascades~\cite{fasterRcnn,cascadeRcnn}, and dense single-stage anchor-based or anchor-free predictors~\cite{yoloOriginal,retinanet,fcos,yoloEleven,yoloTwentySix}. End-to-end query transformers~\cite{detr} eliminate hand-crafted post-processing through bipartite set prediction, with subsequent advances accelerating convergence and real-time throughput via deformable multi-scale sampling, denoising anchors~\cite{dinoDetr}, improved matching~\cite{deim}, and efficient hybrid encoders~\cite{rtDetr}, while knowledge distillation compresses heavy detectors into compact students~\cite{hintonKD,detrDistill,kdDetr}. Under adverse illumination and weather, differentiable image-adaptive filtering~\cite{iaYolo}, dark-scene autoencoding~\cite{maetDark}, and zero-shot day--night adaptation~\cite{daiNet} improve robustness, yet visible-only pipelines remain fundamentally bounded when darkness or dense fog erases reflectance cues.

\subsection{Visible--Infrared Fusion Object Detection}
Multispectral benchmarks~\cite{kaistMultispectral,flir,flirAligned,llvip,m3fd} spurred dual-stream fusion architectures that combine thermal saliency with visible texture. Early and lightweight CNNs refine cross-modal features via cyclic blocks~\cite{flirAligned}, illumination- or uncertainty-aware gating~\cite{mbNet,gaff,uaCmDet}, mono-modality representation enhancement~\cite{m2dLif}, self-modulated or guided cross-attention~\cite{lasfNet,lcafNet}, optimal-transport expert routing~\cite{flexibleFusion}, and probabilistic ensembling~\cite{probEn}. To mitigate cross-sensor misalignment, region alignment~\cite{arCnn} and transformer/state-space architectures model global or deformable correspondence~\cite{cft,icaFusion,msDetr,mmDetr,zpaMdetr,sparseRgbt}. However, missing modalities can degrade multimodal transformers~\cite{missingModalRobustness}, and sensor failures also challenge RGB--thermal object detectors~\cite{deNetFailure}. Recent modality-incomplete approaches explore mixed-patch agnostic training~\cite{mipa}, dynamic routing~\cite{drNet}, and modality-agnostic deformable attention in Scarf-DETR~\cite{scarfDetr}, which applies identical heavy blocks across all scales. In contrast, \method{} introduces asymmetric Resolution-Tiered Fusion (RTF) matched to cross-scale token complexity alongside availability-synchronized Mode-Matched Feature Distillation (MMFD) at zero inference overhead.

\section{Method}
\label{sec:method}

\method{} establishes a unified end-to-end detection paradigm under arbitrary modality availability via two synergistic innovations: \emph{Resolution-Tiered Fusion} (RTF) in the feature space and \emph{Mode-Matched Feature Distillation} (MMFD) in the optimization space. RTF pairs a lightweight spatial modality gate on token-dense $P_3$ with channel-compressed deformable Admoab necks on semantic $P_4$ and $P_5$ maps, while an availability-synchronized full-fusion teacher supervises multi-scale encoder features during training (Figure~\ref{fig:method_architecture}). A single student checkpoint supports paired, visible-only, and infrared-only inference without structural switching or duplicate models.

\subsection{Modality-Adaptive Architecture and Routing}
\label{sec:method_routing}

Let $\mathcal{A}\subseteq\{\rgb,\ir\}$ denote the nonempty set of available modalities for an input sample ($\rgb$: visible RGB, $\ir$: thermal infrared). Rather than inferring missingness from pixel statistics or zero-padding absent channels---which wastes FLOPs and shifts normalization statistics---our pipeline explicitly routes active streams $\{I_m : m\in\mathcal{A}\}$, resized to $640\times640$ in a shared frame.

\method{} comprises a weight-shared multi-scale feature stem (instantiated with a lightweight HGNetv2-B2 backbone for real-time efficiency), modality-shared channel projections, the RTF hierarchy, a cross-scale hybrid encoder, and a four-layer query decoder with 300 object queries. When $\mathcal{A}=\{\rgb,\ir\}$, both views are packed into a single batch, and their features are extracted in one parallel pass; when $\mathcal{A}=\{m\}$, absent views are bypassed at input and the backbone runs solely on $I_m$, halving extraction cost. At each level $\ell\in\{3,4,5\}$, a modality-shared $1\times 1$ convolution with batch normalization projects stage features to 256-channel maps $X_\ell^m$ at $80\times 80$ ($P_3$), $40\times 40$ ($P_4$), and $20\times 20$ ($P_5$).

RTF merges $\{X_\ell^m\}_{m\in\mathcal{A}}$ into unified 256-channel maps $F_\ell$ ($\ell\in\{3,4,5\}$), which enter the hybrid encoder (combining $P_5$ intra-scale self-attention with bidirectional FPN/PAN aggregation) to produce multi-scale features $(S_3, S_4, S_5)$ for the shared query decoder and prediction heads across all modes.

\subsection{Resolution-Tiered Fusion}
\label{sec:method_rtf}

Uniform cross-scale attention overlooks two pyramid asymmetries. First, for a $640\times 640$ input, $P_3$ holds $6{,}400$ tokens per view---$4\times$ $P_4$ ($1{,}600$) and $16\times$ $P_5$ ($400$), or $76.2\%$ of all pyramid tokens---dominating uniform attention latency. Second, $P_3$ primarily encodes localized edges and thermal contours where pixel-wise reliability weighting suffices, whereas $P_4$ and $P_5$ encode semantic object regions where residual cross-sensor parallax benefits from learnable spatial sampling. RTF exploits this hierarchy via a local Spatial Modality Gate at $P_3$ and compact deformable Admoab necks at $P_4/P_5$, sharing parameters across modalities within each scale.

\begin{figure}[t]
\centering
\includegraphics[width=\columnwidth]{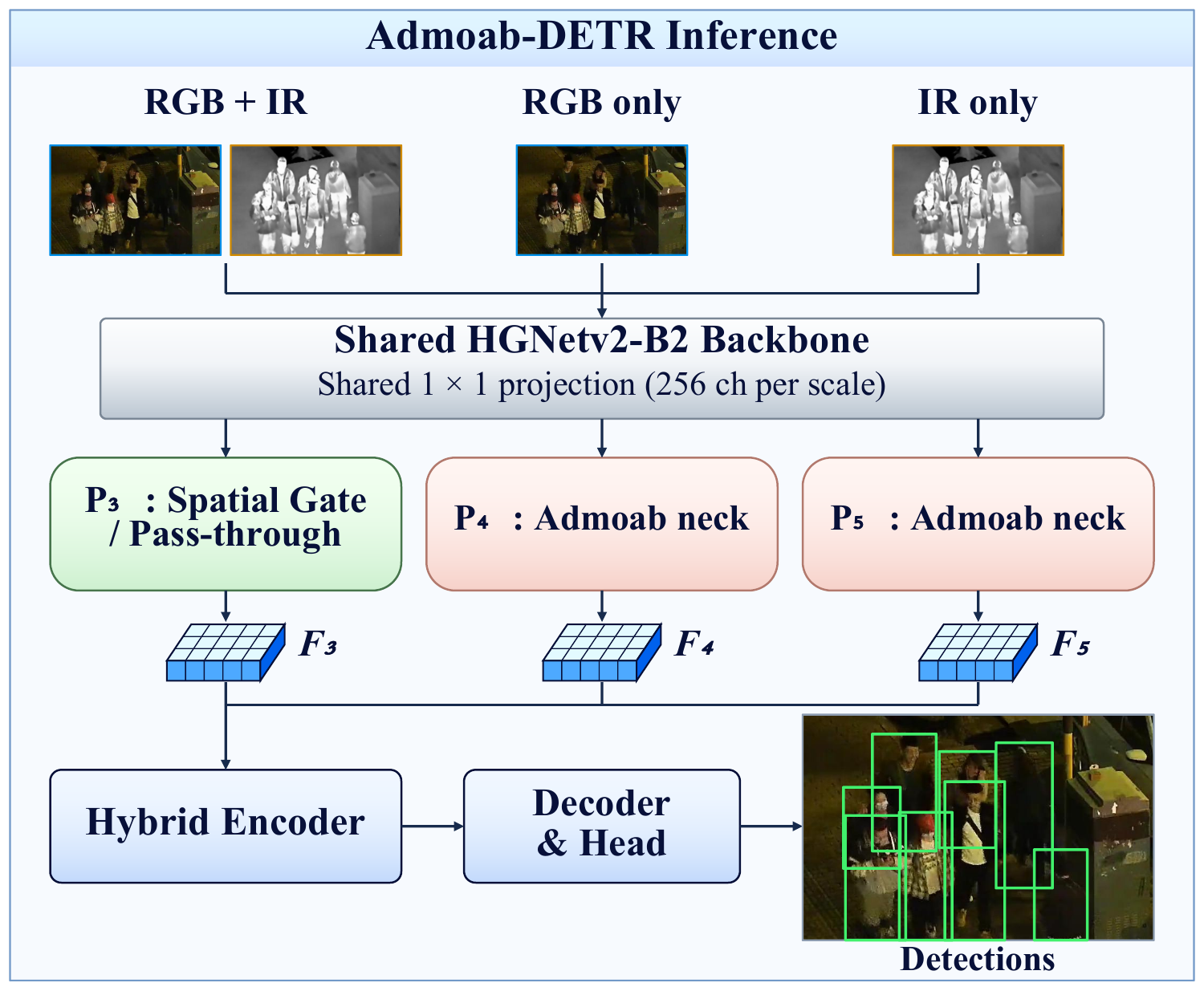}
\caption{\textbf{Dynamic single-checkpoint inference.} Paired or single-view inputs traverse the shared backbone, resolution-tiered fusion, hybrid encoder, and query decoder with zero teacher overhead.}
\label{fig:inference}
\end{figure}

\paragraph{Finest-scale Spatial Modality Gate (P3).}
For paired inputs, the Spatial Modality Gate (SMG, Fig.~\ref{fig:method_architecture}b) concatenates $X_3^\ir$ and $X_3^\rgb$ (512 channels) and predicts local modality reliability at each coordinate $u$ via a $1\times 1$ bottleneck convolution ($512\to 32$, SiLU), a $3\times 3$ depthwise convolution ($32\to 32$, SiLU) capturing neighborhood context, and a $1\times 1$ layer ($32\to 2$). Modality-wise softmax produces weights $w_\ir(u)$ and $w_\rgb(u)$ satisfying $w_\ir(u)+w_\rgb(u)=1$, yielding
\begin{equation}
F_3(u) = w_\ir(u) X_3^\ir(u) + w_\rgb(u) X_3^\rgb(u),
\label{eq:gate}
\end{equation}
where the weights are broadcast over all 256 channels. When $\mathcal{A}=\{m\}$, SMG bypasses gating and routes $F_3 = X_3^m$ at zero fusion cost.

\paragraph{Compact Admoab Neck (P4 and P5).}
At each coarse scale $\ell \in \{4, 5\}$ (Fig.~\ref{fig:method_architecture}c), a bias-free shared $1 \times 1$ layer $C_\ell$ compresses $X_\ell^m$ from $256$ to $d = 128$ channels.
Tokens with 2D sinusoidal positional and learnable scale embeddings traverse one Admoab block equipped with \emph{Admoab Modality-Agnostic Deformable Attention} (A-MADA) ($H = 4$ heads, $K = 4$ points per view).
In MADA~\cite{scarfDetr}, attention logits $a_{q,s,k}$ are predicted solely from the query token, leaving weights blind to whether sampled cross-modal locations fall on misaligned background or degraded regions.
A-MADA retains $a_{q,s,k}$ as a query prior and adds a lightweight content-based calibration term exclusively over the sparse sampled points:
\begin{equation}
\widetilde{a}_{q,s,k} = a_{q,s,k} + \beta\,\operatorname{sim}\!\big(Q_q,\, K_s(p_q + \Delta p_{q,s,k})\big),
\label{eq:amada}
\end{equation}
with normalized attention weights $A = \operatorname{softmax}(\widetilde{a})$ computed across active $(s, k)$ pairs: paired views normalize over $2K = 8$ cross/intra-modal samples, whereas a single view occupies both slots and normalizes jointly over $4K = 16$ samples.
Here, $s$ indexes the source modality and $k$ indexes the sampling point; $Q_q$ and $K_s(p_q + \Delta p_{q,s,k})$ denote low-dimensional projections of query $q$ and the already bilinearly sampled features, reusing sparse samples without dense full-map attention.
Initializing the learnable scalar to $\beta = 0$ ensures that A-MADA initially reproduces the behavior of the original MADA at $t = 0$.

After residual LayerNorm, a residual SwiGLU FFN (width $2d=256$), and a second LayerNorm, concatenated modality outputs $(Y_\ell^\ir, Y_\ell^\rgb)$ are projected via $P_\ell$ ($256\to 256$) onto an uncompressed full-width residual anchor:
\begin{equation}
F_\ell = P_\ell\!\left(\cat(Y_\ell^\ir, Y_\ell^\rgb)\right) + \frac{X_\ell^\ir + X_\ell^\rgb}{2}.
\label{eq:coarse}
\end{equation}
For $\mathcal{A}=\{m\}$, $Y_\ell^m$ fills both slots over an identity anchor, $F_\ell = P_\ell\!\left(\cat(Y_\ell^m, Y_\ell^m)\right) + X_\ell^m$, preserving 256-channel high-frequency fidelity alongside half-width deformable sampling.

\paragraph{Modality-consistent zero-initialized anchoring.}
Zero-initializing the final $1\times 1$ layer of SMG and $P_\ell$ in every Admoab neck enforces an exact boundary condition at $t=0$: uniform softmax gates ($w_\ir=w_\rgb=0.5$) and $P_\ell(\cdot)=0$ reduce paired fusion across all scales to unbiased averaging $(X_\ell^\ir + X_\ell^\rgb)/2$ and single-view fusion to exact identity $X_\ell^m$, preventing early random attention perturbations from destabilizing multi-mode optimization.

\subsection{Mode-Matched Feature Distillation}
\label{sec:method_mmfd}

\paragraph{Isomorphic full-fusion teacher and latent alignment.}
To recover full cross-modal capacity without inference overhead, we train an isomorphic teacher sharing the student's backbone, hybrid encoder, and decoder interfaces while replacing RTF with two full-width Admoab blocks at \emph{every} scale $\ell\in\{3,4,5\}$ ($d=256$, $H=8$, $K=4$, SwiGLU width 512, zero-initialized output projections). The teacher is initialized from standard COCO-pretrained detection weights with its classification heads reinitialized. After convergence, it is frozen, and its non-fusion parameters initialize the student to align latent spaces. The student's RTF modules are initialized separately.

\paragraph{Group-balanced pseudo-modality dropout.}
Both models train with random dropout ~\cite{scarfDetr}: each of $N$ pairs stays intact with probability $1-p$ ($p=0.6$) or splits with probability $p$ into \emph{both} RGB-only and IR-only samples sharing ground-truth boxes. Grouping $n_{\mathrm{p}}$ intact pairs and $2n_{\mathrm{s}}$ split views ($n_{\mathrm{p}}+n_{\mathrm{s}}=N$) with loss weights $n_{\mathrm{p}}/N$ and $n_{\mathrm{s}}/N$ retains all images without double-counting split pairs.

\paragraph{Availability-synchronized multi-scale supervision.}
Unlike cross-modal distillation that forces single-view students to hallucinate paired teacher features, MMFD feeds the frozen teacher the \emph{exact same} dropout samples and availability set $\mathcal{A}$ as the student (Fig.~\ref{fig:method_architecture}a), ensuring targets reflect observable cues. Without running the teacher decoder, we align encoder outputs $(S_3, S_4, S_5)$ to $(T_3, T_4, T_5)$ via channel-wise cosine distance:
\begin{equation}
\begin{aligned}
\mathcal{L}_{\mathrm{KD}}^{g}
&= \frac{1}{3B_g}\sum_{\ell=3}^{5}\sum_{i=1}^{B_g}\frac{1}{|\Omega_\ell|} \\
&\quad\times\sum_{u\in\Omega_\ell}\left[1 - \cos\!\left(S_{\ell,i}(u), \sg(T_{\ell,i}(u))\right)\right],
\end{aligned}
\label{eq:kd}
\end{equation}
for group size $B_g\in\{n_{\mathrm{p}}, 2n_{\mathrm{s}}\}$, grid $\Omega_\ell$, and stop-gradient $\sg(\cdot)$ ($\epsilon=10^{-6}$). Over nonempty dropout groups $\mathcal{G}$, the student minimizes
\begin{equation}
\mathcal{L}_{\mathrm{student}} = \sum_{g\in\mathcal{G}}\frac{n_g}{N}\left(\mathcal{L}_{\mathrm{det}}^{g} + \lambda(t)\mathcal{L}_{\mathrm{KD}}^{g}\right),
\label{eq:student_objective}
\end{equation}
with detection loss $\mathcal{L}_{\mathrm{det}}^{g}$ and a five-epoch linear warmup $\lambda(t)=\min(1, t/5)$. Only the compact student is deployed (Figure~\ref{fig:inference}).

\section{Experiments}
\label{sec:experiments}

\subsection{Admoab Datasets}
\begin{figure}[t]
  \centering
  \includegraphics[width=\columnwidth]{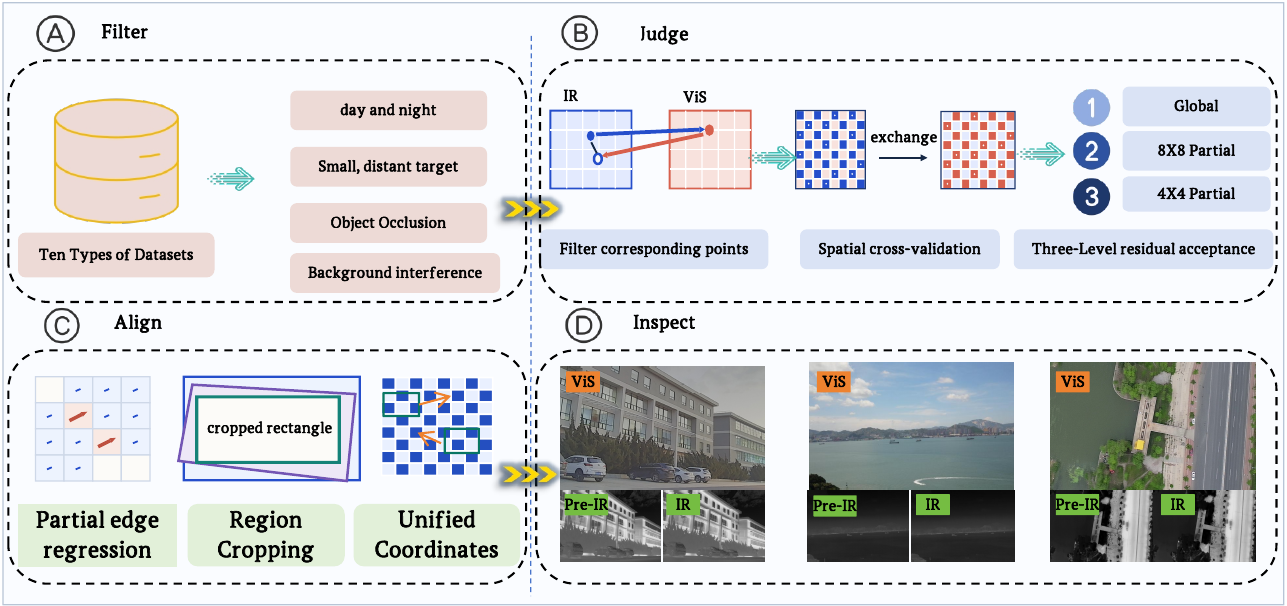}
  \caption{\textbf{Admoab dataset construction pipeline.} Four-stage workflow of dataset filtering, alignment judgment, geometric registration, and consistency inspection.}
  \label{fig:dataset_pipeline}
  \label{fig:runtime}
\end{figure}

\paragraph{Dataset Filtering.}
We compiled 67,168 labeled image pairs from nine publicly available datasets: LLVIP~\cite{llvip}, M3FD~\cite{m3fd}, FLIR-Aligned~\cite{flir,flirAligned}, AIRST-CS~\cite{airstcs}, RTDOD~\cite{rtdod}, UAV-MSR~\cite{uavmsr}, VTUAV-det~\cite{vtuav,rgbt_drone}, RGBTDronePerson~\cite{rgbt_drone}, and SEA-AI SEADs~\cite{sea_ai}, supplemented with 3,256 self-collected pairs covering ground vehicles, pedestrians, maritime vessels, drones, and wildlife scenes. Using visible light as the reference view, we assessed cross-modal alignment across all 70,424 candidate image pairs. Figure~\ref{fig:dataset_pipeline} illustrates the four-stage dataset construction workflow: dataset filtering, alignment judgment, geometric registration, and consistency inspection.

\begin{figure}[t]
  \centering
  \includegraphics[width=\columnwidth]{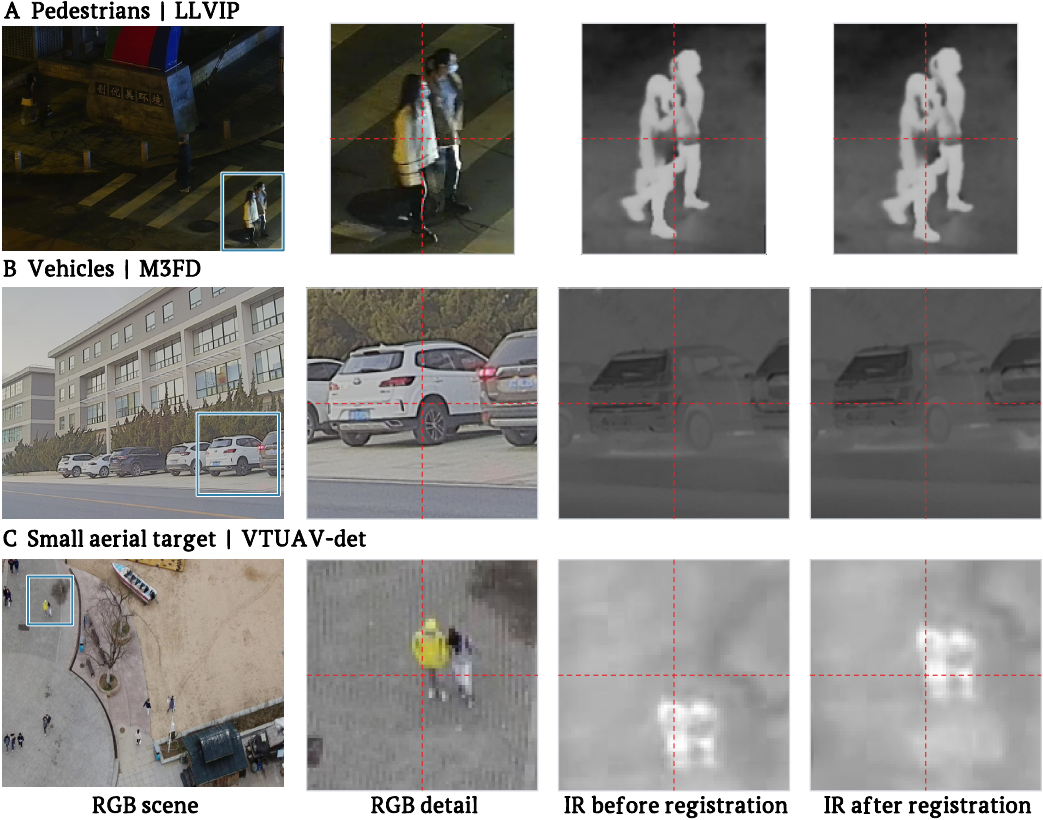}
  \caption{\textbf{Visualization before and after registration} across representative scenes from LLVIP, M3FD, and VTUAV-det.}
  \label{fig:sixfig}
\end{figure}

\paragraph{Data Alignment and Testing.}
The alignment assessment uses the visible-light image as the reference and extracts cross-modal correspondences with RoMa~\cite{roma}, using the RoMa weights released with AnyMatch~\cite{anymatch}, a multimodal training-data synthesis framework. The correspondences are filtered by a bidirectional consistency check. A reference mapping is directly retained if at least 90\% of matched points exhibit an error within 3 pixels, the P90 error is at most 2 pixels, and local residual checks pass. For the remaining pairs with sufficient matches, affine and homography transformations are fitted, validated via two-fold cross-validation on an $8\times 8$ grid, and subsequently refitted on all inliers. After geometric verification, crop and annotation checks, and deduplication, the initial cohort contained 13,507 pairs. We supplemented this cohort with additional self-collected data. Figure~\ref{fig:sixfig} visualizes representative scenes from three datasets before and after registration, demonstrating substantially improved spatial alignment.

\paragraph{Dataset Partitioning.}
 To reduce leakage between correlated samples, we split by source and provenance group, keeping recorded sequences, capture batches, and adjacent-frame groups intact. We target a 70\%/15\%/15\% training/validation/test ratio to retain most data for learning while reserving separate sets for model selection and evaluation. The resulting split contains 69.76\% of the image pairs for training, 15.76\% for validation, and 14.48\% for testing, respectively.  Mixed-integer optimization minimizes normalized deviations from these targets in total image count, source image counts, class box counts, and joint source--class box counts. Both registration conditions use identical split membership, visible images, and annotations. Sources and classes spanning at least three groups are required to appear in all splits.

\begin{table*}[!t]
  \caption{\textbf{Detection on M3FD~\cite{m3fd}, FLIR-aligned~\cite{flirAligned}, and LLVIP~\cite{llvip}.} A comparison of object detection methods using single-modality inputs (infrared or visible light) versus paired infrared--visible inputs.}
  \label{tab:performance}
  \centering
  \small
  \setlength{\tabcolsep}{3pt}
  \renewcommand{\arraystretch}{0.96}
  \begin{tabular*}{\textwidth}{@{\extracolsep{\fill}}lcrrrrrrrrr@{}}
    \toprule
    & & \multicolumn{3}{c}{\textbf{M3FD}}
    & \multicolumn{3}{c}{\textbf{FLIR-aligned}}
    & \multicolumn{3}{c}{\textbf{LLVIP}} \\
    \cmidrule(lr){3-5}\cmidrule(lr){6-8}\cmidrule(l){9-11}
    \textbf{Method} & \textbf{Input} & \textbf{mAP} & \textbf{AP$_{50}$} & \textbf{AP$_{75}$}
    & \textbf{mAP} & \textbf{AP$_{50}$} & \textbf{AP$_{75}$}
    & \textbf{mAP} & \textbf{AP$_{50}$} & \textbf{AP$_{75}$} \\
    \midrule
    YOLO11m~\cite{yoloEleven} & ViS & 39.02 & 56.08 & 42.96 & 27.34 & 59.79 & 21.46 & 46.48 & 81.61 & 47.39 \\
    YOLO26m~\cite{yoloTwentySix} & ViS & 39.12 & 56.52 & 42.98 & 26.57 & 59.16 & 19.88 & 47.98 & 83.63 & 50.37 \\
    RT-DETRv2-S/R18~\cite{rtDetrTwo} & ViS & 34.04 & 50.86 & 36.55 & 25.98 & 58.48 & 19.52 & 43.01 & 77.90 & 42.82 \\
    RT-DETRv4-M~\cite{rtdetrv4} & ViS & 39.91 & 56.86 & 43.30 & 27.15 & 60.47 & 20.91 & 47.49 & 83.86 & 48.50 \\
    DEIM-RT-DETR~\cite{deim} & ViS & 40.90 & 65.65 & 41.70 & 34.52 & 69.37 & 28.45 & 53.99 & 91.79 & 58.40 \\
    CFT~\cite{cft} & ViS & 34.5 & 59.4 & 34.4 & 23.6 & 55.7 & 16.6 & 21.8 & 49.0 & 16.2 \\
    ICAFusion~\cite{icaFusion} & ViS & 25.0 & 48.6 & 23.0 & 24.4 & 60.3 & 16.5 & 1.0 & 2.7 & 0.0 \\
    DAMS-DETR~\cite{damsDet} & ViS & 46.6 & 71.6 & 48.8 & 28.4 & 61.4 & 22.4 & 15.0 & 38.1 & 8.7 \\
    Scarf-Align-DETR~\cite{scarfDetr} & ViS & 51.4 & 77.5 & 54.9 & 37.8 & 76.5 & 32.3 & 55.5 & 92.6 & 60.3 \\
    \textbf{Admoab-DETR} & \textbf{ViS} & \textbf{52.71} & \textbf{76.21} & \textbf{57.61} & \textbf{39.68} & \textbf{77.54} & \textbf{34.82} & \textbf{57.52} & \textbf{93.09} & \textbf{63.58} \\
    \midrule
    YOLO11m~\cite{yoloEleven} & IR & 16.33 & 27.11 & 16.90 & 40.03 & 71.89 & 38.15 & 54.14 & 81.42 & 60.28 \\
    YOLO26m~\cite{yoloTwentySix} & IR & 19.01 & 31.90 & 19.39 & 39.97 & 71.60 & 38.15 & 55.15 & 80.70 & 62.61 \\
    RT-DETRv2-S/R18~\cite{rtDetrTwo} & IR & 15.71 & 26.99 & 15.44 & 36.24 & 67.71 & 32.94 & 46.30 & 74.20 & 49.47 \\
    RT-DETRv4-M~\cite{rtdetrv4} & IR & 19.37 & 33.21 & 19.39 & 39.21 & 70.92 & 36.25 & 54.01 & 81.89 & 59.64 \\
    DEIM-RT-DETR~\cite{deim} & IR & 37.52 & 60.35 & 39.42 & 45.75 & 81.73 & 42.83 & 67.90 & 97.54 & 80.50 \\
    CFT~\cite{cft} & IR & 12.6 & 23.8 & 12.4 & 31.0 & 63.3 & 24.8 & 60.4 & 94.0 & 66.2 \\
    ICAFusion~\cite{icaFusion} & IR & 20.3 & 39.5 & 19.0 & 34.8 & 72.9 & 27.9 & 63.4 & 97.2 & 73.1 \\
    DAMS-DETR~\cite{damsDet} & IR & 26.9 & 44.4 & 27.5 & 47.0 & 82.6 & 45.5 & 69.3 & 97.6 & 78.3 \\
    Scarf-Align-DETR~\cite{scarfDetr} & IR & 41.0 & 65.2 & 42.0 & 47.3 & 83.1 & 45.4 & 69.3 & 98.2 & 82.2 \\
    \textbf{Admoab-DETR} & \textbf{IR} & \textbf{43.32} & \textbf{67.26} & \textbf{44.97} & \textbf{49.70} & \textbf{84.46} & \textbf{49.91} & \textbf{70.73} & \textbf{98.32} & \textbf{82.42} \\
    \midrule
    CFT~\cite{cft} & ViS + IR & 42.5 & 68.2 & 44.6 & 40.2 & 78.7 & 35.5 & 63.6 & 97.5 & 72.9 \\
    DAMS-DETR~\cite{damsDet} & ViS + IR & 52.9 & 80.2 & 56.0 & 49.3 & 86.6 & 48.1 & 69.6 & 97.9 & 79.1 \\
    Scarf-Align-DETR~\cite{scarfDetr} & ViS + IR & 54.4 & 80.9 & 57.5 & 49.6 & 85.4 & 48.9 & 69.6 & 98.4 & 82.9 \\
    DeformCAT~\cite{deformCat} & ViS + IR & 46.49 & 73.28 & 49.78 & 46.93 & 86.48 & 43.65 & 66.13 & 97.82 & 77.07 \\
    GM-DETR~\cite{gmDetr} & ViS + IR & 45.72 & 71.25 & 47.15 & 45.80 & 83.90 & 42.60 & 70.2 & 97.4 & 81.4 \\
    ZPA-MDETR ($Q=\mathrm{IR}$)~\cite{zpaMdetr} & ViS + IR & 50.58 & 77.26 & 53.18 & 48.16 & 86.65 & 44.89 & 64.88 & 97.82 & 76.26 \\
    ZPA-MDETR ($Q=\mathrm{ViS}$)~\cite{zpaMdetr} & ViS + IR & 48.39 & 74.36 & 50.68 & 49.82 & 87.05 & 48.32 & 69.14 & 98.11 & 78.79 \\
    \addlinespace[2pt]
\textbf{Admoab-DETR} & \textbf{ViS + IR} & \textbf{55.65} & \textbf{79.92} & \textbf{60.30} & \textbf{50.99} & \textbf{86.67} & \textbf{51.64} & \textbf{70.95} & \textbf{98.42} & \textbf{82.94} \\
\textbf{Admoab-DETR-X} & \textbf{ViS + IR} & \textbf{59.64} & \textbf{85.20} & \textbf{64.24} & \textbf{51.41} & \textbf{87.95} & \textbf{51.88} & \textbf{71.39} & \textbf{98.59} & \textbf{84.22} \\
    \bottomrule
  \end{tabular*}
\end{table*}

\subsection{Implementation Details}
\label{sec:implementation_details}

Admoab-DETR utilizes a shared HGNetv2-B2 backbone and inherits unfused weights from a frozen fully-fused teacher model. The model is trained using the AdamW optimizer with a global batch size of 16 image pairs and a weight decay of $10^{-4}$. The initial learning rates are set to $10^{-4}$ for the fusion and classification layers, and $10^{-5}$ for the remaining parts. By default, random Dropout is applied with a probability of $p=0.6$, and knowledge distillation is performed on all three outputs of the hybrid encoder. We report COCO-style mAP, $\text{AP}_{50}$, and $\text{AP}_{75}$~\cite{coco}, alongside parameter count, GFLOPs, and GPU inference speed.

Object detection performance is evaluated on three benchmark datasets: M3FD~\cite{m3fd} (containing 831 test pairs of visible-infrared traffic scenes under diverse lighting and weather conditions), FLIR-aligned~\cite{flir,flirAligned} (featuring 1,013 registered pairs annotated with pedestrians, cars, and bicycles), and LLVIP~\cite{llvip} (focusing on pedestrian detection in low-light environments with 3,463 pairs).

\subsection{Experiments and Results Evaluation}
\label{sec:input_modes}
\label{sec:fusion_capacity}

\paragraph{Quantitative Comparison.}
To evaluate the effectiveness of Admoab-DETR across different input modalities, we conduct comprehensive experiments on three multimodal benchmarks: M3FD, FLIR-aligned, and LLVIP. Table~\ref{tab:performance} presents the quantitative comparison, where the best results per modality are highlighted in bold. The experimental results demonstrate that, compared with existing unimodal (infrared or visible) and multimodal (infrared-visible) detectors, our approach consistently achieves superior performance across all datasets and modality configurations. Furthermore, scaling up to a larger variant, Admoab-DETR-X, yields further accuracy gains, validating the effectiveness and superiority of our method across diverse modalities.

\begin{figure*}[!t]
\centering
\includegraphics[width=\textwidth]{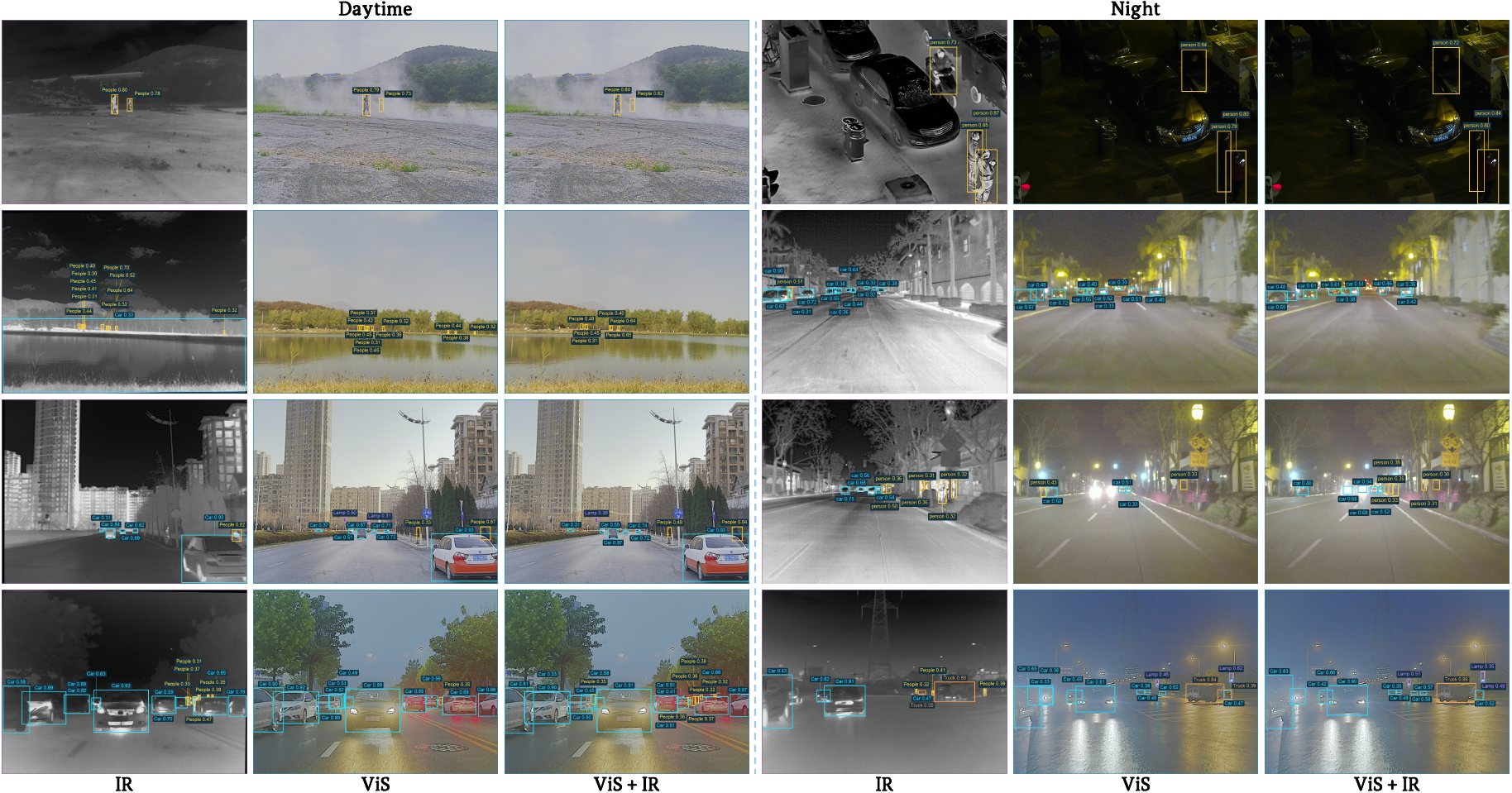}
  \caption{\textbf{Qualitative comparison across input modes.} Qualitative comparison of object detection results across different input modalities (ViS, IR, and ViS+IR) on the M3FD, FLIR-aligned, and LLVIP datasets. ViS, IR, and ViS+IR denote visible, infrared, and fused modalities, respectively. Our method demonstrates superior robustness and accuracy under challenging conditions.}
  \label{fig:admoab_detection_examples}
\end{figure*}

\paragraph{Qualitative Analysis.}
Figure~\ref{fig:admoab_detection_examples} compares detection results across various environmental conditions. Under adequate illumination, the visible modality preserves fine textures and appearance features vital for target discrimination. Conversely, in night-time or low-light scenarios, where thermal radiation becomes the key cue, the infrared modality alone often yields more reliable detection. In challenging scenarios involving distant small objects, severe occlusion, and complex backgrounds, leveraging cross-modal complementarity allows bimodal input to recover otherwise missed targets. Although single-modality performance is influenced by distance, lighting, and occlusion, real-world scenes typically present a combination of these challenges. Therefore, for robust all-weather deployment, a unified detection framework must support both unimodal and bimodal processing paths, ensuring continuous operation during sensor degradation while reducing processing latency when unimodal data is sufficient.

\begin{table}[!htbp]
 \caption{\textbf{Model Efficiency.} Comparison of parameters, GFLOPs, and latency on an H800 at $640 \times 640$ input.}
  \label{tab:admoab_efficiency}
  \centering
  \footnotesize
  \setlength{\tabcolsep}{1.5pt}
  \begin{tabular*}{\columnwidth}{@{\extracolsep{\fill}}lcrrr@{}}
    \toprule
    \textbf{Method} & \textbf{Input} & \shortstack{\textbf{Params}\\\textbf{(M)}} & \textbf{GFLOPs} & \shortstack{\textbf{Latency}\\\textbf{(ms)}} \\
    \midrule
    YOLO26m & ViS & 20.411 & 68.4 & 1.028 \\
    RT-DETRv2 & ViS & 74.773 & 251.7 & 1.909 \\
    RT-DETRv4 & ViS & 61.696 & 203.1 & 2.289 \\
    \textbf{Admoab-DETR} & \textbf{ViS/IR} & \textbf{19.753} & \textbf{57.951} & \textbf{1.587} \\
    \midrule
    DAMSDet & ViS+IR & 79.062 & 270.3 & 2.595 \\
    GM-DETR & ViS+IR & 70.362 & 345.7 & 3.824 \\
    Scarf-Align & ViS+IR & 322.552 & 2384.5 & 16.042 \\
    \textbf{Admoab-DETR} & \textbf{ViS+IR} & \textbf{19.753} & \textbf{79.4} & \textbf{1.730} \\
    \bottomrule
  \end{tabular*}

\end{table}

\paragraph{Model Efficiency.}
Table~\ref{tab:admoab_efficiency} compares Admoab-DETR with existing state-of-the-art unimodal and visible-infrared detectors in terms of parameter count, computational complexity (GFLOPs), and inference latency. Across both real-time unimodal detectors and multimodal fusion methods, Admoab-DETR achieves superior detection accuracy with the fewest parameters (19.75M), while requiring the lowest computational cost (79.4 GFLOPs) among multimodal counterparts—significantly outperforming fusion-based approaches such as DAMSDet~\cite{damsDet}, GM-DETR~\cite{gmDetr}, and Scarf-Align~\cite{scarfDetr}. Furthermore, for unimodal inputs (visible or infrared), dynamic routing and $P_3$ gate pass-through reduce the computational cost to 58.0 GFLOPs and accelerate inference latency to 1.587 ms. Moreover, our scaled-up variant, Admoab-DETR-X, delivers further accuracy gains across challenging benchmarks.

\begin{table}[!htbp]
  \caption{\textbf{Detection on the Admoab test set before and after registration.} Scores are reported as percentages.}
  \label{tab:admoab_registration_results}
  \centering
  \footnotesize
  \setlength{\tabcolsep}{2pt}
  \begin{tabular*}{\columnwidth}{@{\extracolsep{\fill}}lrrrrrr@{}}
    \toprule
    & \multicolumn{3}{c}{\textbf{Before}}
    & \multicolumn{3}{c}{\textbf{After}} \\
    \cmidrule(lr){2-4}\cmidrule(l){5-7}
    \textbf{Method} & \textbf{mAP} & \textbf{AP$_{50}$} & \textbf{AP$_{75}$}
                      & \textbf{mAP} & \textbf{AP$_{50}$} & \textbf{AP$_{75}$} \\
    \midrule
    DAMSDet-R50~\cite{damsDet}            & 29.49 & 48.47 & 25.19 & 34.34 & 54.04 & 36.48 \\
    GM-DETR-X~\cite{gmDetr}              & 29.76 & 50.49 & 33.67 & 32.35 & 53.82 & 33.89 \\
    LCAFNet~\cite{lcafNet}               & 11.69 & 24.67 &  9.10 & 14.32 & 31.73 & 11.49 \\
    CFT-L~\cite{cft}                     & 13.90 & 28.18 & 12.28 & 15.95 & 33.25 & 13.07 \\
    ICAFusion-L~\cite{icaFusion}         & 13.98 & 30.35 &  9.97 & 16.17 & 33.64 & 12.84 \\
    Scarf-Align-DETR~\cite{scarfDetr}  & 22.54 & 37.99 & 22.06 & 25.32 & 42.59 & 24.92 \\
    \midrule
    \textbf{Admoab-DETR}                       & \textbf{31.13} & \textbf{50.67} & \textbf{34.06}
                                        & \textbf{36.53} & \textbf{56.67} & \textbf{38.42} \\
    \bottomrule
  \end{tabular*}
\end{table}

\paragraph{Admoab Dataset Alignment Test.}
To validate the effectiveness of our aligned multimodal object detection dataset (Admoab) in boosting detection performance, we conduct comprehensive multispectral detection experiments. Table~\ref{tab:admoab_registration_results} evaluates infrared-visible detectors on the Admoab test set before and after spatial registration. The experimental results demonstrate that, compared with unaligned data, registration consistently yields substantial accuracy gains across all evaluated models. Notably, Admoab-DETR achieves superior performance both prior to registration (31.13 mAP) and post-registration (36.53 mAP, representing a notable improvement of 5.40 points). Featuring diverse all-weather complex scenarios, this dataset establishes a new robust benchmarking platform for multimodal object detection and flexible unimodal-multimodal detection research.

\subsection{Ablation Study}
\label{sec:supervision}
We conduct ablation studies on the FLIR-aligned dataset, investigating feature distillation, random dropout, and cross-modal fusion designs. The detailed ablation results are presented in Table~\ref{tab:admoab_ablation} and Table~\ref{tab:admoab_fusion_ablation}.

\begin{table}[!htbp]
  \caption{\textbf{Distillation and dropout ablations on FLIR-aligned.} Detection results for infrared, visible light, and the fused modality of both.}
  \label{tab:admoab_ablation}
  \centering
  \footnotesize
  \setlength{\tabcolsep}{2pt}
  \begin{tabular*}{\columnwidth}{@{\extracolsep{\fill}}lrrr@{}}
    \toprule
    & \multicolumn{3}{c}{\textbf{mAP (\%)}} \\
    \cmidrule(l){2-4}
    \textbf{Setting} & \textbf{ViS+IR} & \textbf{ViS} & \textbf{IR} \\
    \midrule
    \multicolumn{4}{@{}l}{\textit{Distillation}} \\
    No distillation        & 48.21 & 38.33 & 47.56 \\
    Coarsest scale only    & 49.97 & 39.01 & 48.23 \\
    Full-view teacher      & 48.99 & 38.92 & 47.91 \\
\textbf{MMFD (3 scales)} & \textbf{50.99} & \textbf{39.68} & \textbf{49.70} \\
    \midrule
    \multicolumn{4}{@{}l}{\textit{Dropout probability}} \\
    0.0                    & 49.83 & 39.02 & 48.85 \\
    0.3                    & 50.27 & 39.17 & 49.31 \\
\textbf{0.6} & \textbf{50.99} & \textbf{39.68} & \textbf{49.70} \\
    0.9                    & 50.21 & 39.14 & 48.99 \\
    \bottomrule
  \end{tabular*}
\end{table}

\paragraph{Distillation.}
As shown in Rows 1--4 of Table~\ref{tab:admoab_ablation}, our three-scale MMFD boosts mAP by 2.78, 1.35, and 2.14 points on ViS+IR, ViS, and IR over the baseline. It also surpasses coarsest-scale supervision and full-view teacher guidance, confirming that modality-matched targets mitigate cross-modal hallucination.

\paragraph{Dropout probability.}
The results of adjusting the dropout probability are shown in rows 5–8 of Table~\ref{tab:admoab_ablation}; a value of $p=0.6$ achieves the optimal multimodal balance, whereas an excessively high dropout probability leads to insufficient training of the paired fusion component.
\begin{table}[!t]
  \caption{\textbf{Fusion strategies.} Latency is measured with paired ViS+IR input on an H800.}
  \label{tab:admoab_fusion_ablation}
  \centering
  \footnotesize
  \setlength{\tabcolsep}{3pt}
  \begin{tabular*}{\columnwidth}{@{\extracolsep{\fill}}lrrrrr@{}}
    \toprule
    & \multicolumn{3}{c}{\textbf{mAP (\%)}}
    & \multicolumn{2}{c}{\textbf{ViS + IR}} \\
    \cmidrule(lr){2-4}\cmidrule(l){5-6}
    \textbf{Setting} & \textbf{ViS+IR} & \textbf{ViS} & \textbf{IR}
    & \shortstack{\textbf{Params (M)}} & \textbf{ms} \\
    \midrule
    Mean fusion              & 48.87 & 38.75 & 47.80 & 19.1916 & 1.570 \\
    Full fusion  & 50.74 & 39.33 & 49.46 & 24.1023 & 2.867 \\
    \shortstack[l]{Uniform lightweight fusion} & 49.67 & 38.21 & 48.53 & 20.0136 & 1.946 \\
    \textbf{RTF}             & \textbf{50.99} & \textbf{39.68} & \textbf{49.70} & \textbf{19.7508} & \textbf{1.731} \\
    \bottomrule
  \end{tabular*}
\end{table}
\paragraph{Fusion strategy.}
As shown in Table~\ref{tab:admoab_fusion_ablation}, resolution-tiered fusion improves mAP across all modalities over uniform lightweight fusion while reducing paired latency, validating our asymmetric scale allocation. Furthermore, RTF surpasses full fusion in accuracy while saving 18.1\% parameters and 39.6\% paired latency.

\FloatBarrier

\section{Conclusion}
\label{sec:conclusion}
We presented \method{}, a unified real-time detector for infrared--visible object detection under arbitrary modality availability. Resolution-Tiered Fusion (RTF) decouples fine-scale pixel-level modality gating with zero-cost single-view pass-through from coarse-scale channel-compressed deformable aggregation, overcoming the latency bottleneck of uniform cross-scale attention, while Mode-Matched Feature Distillation (MMFD) transfers full-fusion capacity under synchronized modality availability without cross-modal hallucination or inference overhead. Supported by our strictly registered 20,000-pair Admoab benchmark, \method{} achieves state-of-the-art accuracy and real-time speed across paired, visible-only, and infrared-only regimes within one checkpoint.

\section*{LLM Usage Statement}
Large Language Models (OpenAI Codex, OpenAI GPT, and Google Gemini) were used to assist with code implementation, training and evaluation script preparation, pipeline automation, and manuscript editing for grammar and clarity. All reported quantitative and qualitative results were generated from actual model training and inference runs on physical hardware, and the authors have thoroughly verified and take full responsibility for all code, experiments, and text in this paper.

{\small
\bibliographystyle{ieeenat_fullname}
\bibliography{main}
}

\end{document}